\documentclass[conference]{IEEEtran}
\IEEEoverridecommandlockouts

\usepackage{multirow}
\usepackage{amsfonts}
\usepackage{bm}
\usepackage{enumitem}

\usepackage{booktabs}

\usepackage{amsmath}
\usepackage{amssymb}
\usepackage{mathtools}
\usepackage{caption}

\usepackage{subcaption}
\usepackage{subfiles}

\usepackage{cite}
\usepackage{algorithm}
\usepackage{algorithmic}
\usepackage{graphicx}
\usepackage{textcomp}
\usepackage{xcolor}
\usepackage{soul}
\def\BibTeX{{\rm B\kern-.05em{\sc i\kern-.025em b}\kern-.08em
    T\kern-.1667em\lower.7ex\hbox{E}\kern-.125emX}}
\begin{document}

\title{Adaptive von Mises–Fisher Likelihood Loss for Supervised Deep Time Series Hashing}

\author{
    \IEEEauthorblockN{Juan Manuel Perez\IEEEauthorrefmark{1}, 
                      Kevin Garcia\IEEEauthorrefmark{1}, 
                      Brooklyn Berry\IEEEauthorrefmark{1}, 
                      Dongjin Song\IEEEauthorrefmark{2}, 
                      Yifeng Gao\IEEEauthorrefmark{1}}
    \IEEEauthorblockA{\IEEEauthorrefmark{1}The University of Texas Rio Grande Valley, Edinburg, TX, USA \\
    Emails: juan.m.perez02@utrgv.edu, kevin.garcia09@utrgv.edu, brooklyn.berry01@utrgv.edu, yifeng.gao@utrgv.edu}
    \IEEEauthorblockA{\IEEEauthorrefmark{2}University of Connecticut, Storrs, CT, USA\\
    Email: dongjin.song@uconn.edu}
}

\maketitle

\begin{abstract}

Indexing time series by creating compact binary representations is a fundamental task in time series data mining. Recently, deep learning-based hashing methods have proven effective for indexing time series based on semantic meaning rather than just raw similarity. The purpose of deep hashing is to map samples with the same semantic meaning to identical binary hash codes, enabling more efficient search and retrieval. Unlike other supervised representation learning methods, supervised deep hashing requires a discretization step to convert real-valued representations into binary codes, but this can induce significant information loss. In this paper, we propose a von Mises–Fisher (vMF) hashing loss. The proposed deep hashing model maps data to an $M$-dimensional hyperspherical space to effectively reduce information loss and models each data class as points following distinct vMF distributions. The designed loss aims to maximize the separation between each modeled vMF distribution to provide a better way to maximize the margin between each semantically different data sample. Experimental results show that our method outperforms existing baselines. The implementation is publicly available at https://github.com/jmpq97/vmf-hashing


\end{abstract}

\section{Introduction}

Time series data consists of sequential measurements collected over time and is commonly used to analyze trends, identify patterns, and forecast future events \cite{camerra2010isax,palpanas2020evolution,yagoubi2017dpisax, lin2007experiencing,zhang2020semantic,garcia2024efficient,feng2024efficient}. It plays an important role in a wide range of fields such as finance \cite{tsay2005analysis,sezer2020financial, kim2003financial}, healthcare \cite{chuah2007ecg,pyakillya2017deep}, meteorology \cite{hewage2020temporal, karevan2020transductive}, and energy \cite{sun2024entropy, alvarez2010energy}, as it supports making informed decisions based on historical data.

The high dimensionality and often continuous, streaming nature of time series data make brute-force search computationally infeasible. Therefore, a reliable indexing method is essential to organize data to enable fast similarity search, avoiding the need to scan each time series individually. 

Traditional indexing methods, such as iSAX (indexable Symbolic Aggregate approXimation) \cite{shieh2008sax,camerra2010isax,palpanas2019evolution}, rely on symbolic representation and dimensionality reduction to enable fast similarity search and scalability for large time series datasets. While such indexing approaches are efficient, they are often designed to reflect specific distance measures such as Euclidean distance or Dynamic Time Warping (DTW). As a result, they struggle to capture similarity at the semantic level. On the other hand, indexing data based on its semantic meaning captured by supervised deep models has become increasingly popular. These models enable the retrieval of meaningful behaviors (e.g., walking, running, or cycling) that are useful for a wide range of downstream tasks. Deep hashing models \cite{hoe2021one, cao2017hashnet, su2018greedy, zhu2016deep} learn binary embeddings that map time series of the same category close together. Compared to traditional approaches, they generally offer a better semantically aligned similarity measure.

\begin{figure*}[t]
     \centering
         \includegraphics[width=1\textwidth]{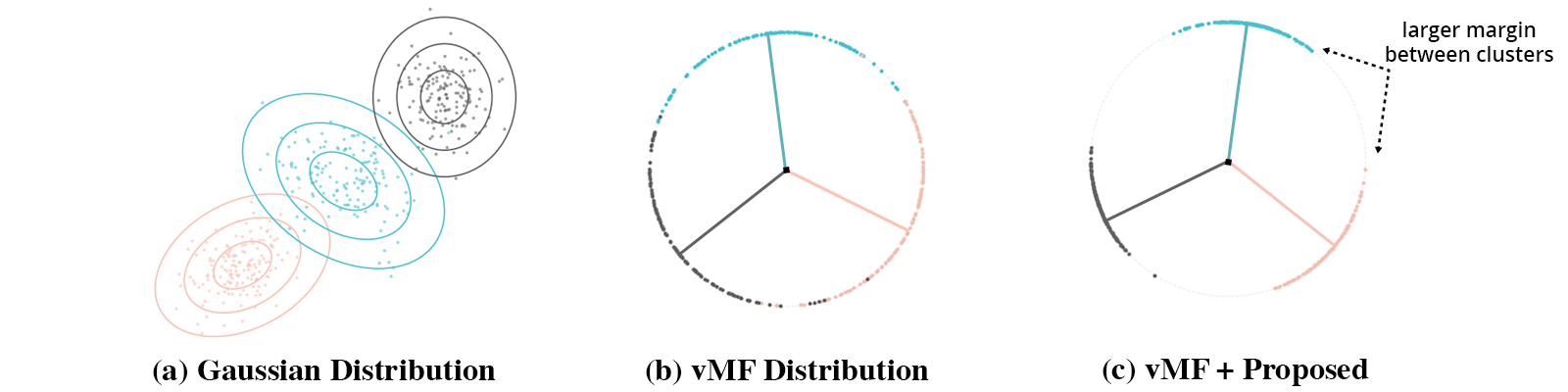}
         \caption{(a) Clusters in $\mathbb{R}^2$, (b) Clusters modeled via vMF prior in $\mathcal{S}^1$, and (c) Margin-aware vMF to clearly separate class clusters.}
         \label{fig:proposed}
\end{figure*}

Despite significant advances, the discretization operation, which converts real-valued embeddings into binary code via $\mathrm{sign}(x)=\mathbf{1}\{x>0\}$, may lead to significant information loss and can dramatically reduce model performance. The intuition is illustrated in Figure~\ref{fig:proposed}. Traditional real-valued embeddings learned from deep learning models often fall in $\mathbb{R}^d$ spaces (Figure \ref{fig:proposed}.a). In this case, $\mathrm{sign}(x)$ may incur significant information loss. To tackle the challenge above, we propose a modified von Mises–Fisher (vMF)-based approach to reduce the information loss. Specifically, our method contains two components: we first map embeddings to the hypersphere via a vMF-based prototypical loss (Figure \ref{fig:proposed}.b). Next, during training, we enforce an adaptive margin method by penalizing samples that remain within the margin (Figure \ref{fig:proposed}.c). This encourages compact intra-class grouping and better inter-class separation, reducing ambiguity during retrieval. As a result, semantic structure is preserved even after binarization, and dissimilar sequences are less likely to be retrieved together. The experiments demonstrate that the proposed approach yields less information loss and outperforms several deep hashing baselines.

The rest of the paper is structured as follows. Section II overviews recent deep hashing methods and their challenges with time series data. Section III introduces the problem of learning binary representations and outlines our approach to preserve semantic similarity. In Section IV, we introduce our proposed method, followed by the experiments in Section V. Finally, we conclude the paper in Section VI.

\section{Related Work}

A common bottleneck in today's methods for nearest-neighbor search involves dealing with high-dimensional data and preserving semantic similarity \cite{cao2017hashnet, su2018greedy,zhu2016deep}, particularly in the context of large-scale data retrieval and similarity search tasks. The nearest neighbor problem involves finding the closest or most similar data points to a given query point in a dataset. This problem is central for many applications such as recommendation systems \cite{zhang2020deep}, image retrieval \cite{cao2017hashnet}, and classification tasks \cite{li2017deep}. 

Deep hashing has recently been proposed to reduce data dimensionality while preserving semantic information by learning binary hash codes that capture the underlying meaning of the data. A large number of deep hashing methods have been proposed in the image domain \cite{hoe2021one,su2018greedy,zhu2016deep,cao2017hashnet}. However, all of them are designed for image hashing and the backbones used in these models are 2D convolutional neural networks \cite{krizhevsky2012imagenet,he2016deep}. These models are effective on image datasets, but this presents an issue for time series data.

In contrast to the image domain, very little research has been done in designing deep hashing models for time series data \cite{song2018deep,wang2023seanet}. Song et al. introduced a deep $r^{th}$ root-based loss function to transform time series into binary embeddings for the task of information retrieval. Wang et al. introduced a framework called SeaNet for unsupervised deep hashing with the guidance of symbolic representation. However, none of them address the information loss caused by discretization.

\section{Problem Description}

Given $N$ multivariate time series $x_i\in\mathbb{R}^{D\times T}$ collected into $\mathcal{X}\in\mathbb{R}^{N\times D\times T}$, with labels $y_i\in\{1,\dots,C\}$, the goal of deep hashing is to train a model $f:\mathbb{R}^{D\times T}\to\{-1,+1\}^M$ that maps each series to a binary code $b_i=f(x_i)\in\{-1,+1\}^M$. We desire $b_i$ and $b_j$ to be identical or similar (under Hamming distance) when $y_i=y_j$, and far apart otherwise. Such binary codes enable efficient indexing and similarity search.

\section{Methodology}

\subsection{von Mises–Fisher (vMF) Distribution}

\begin{figure}[htbp]
    \centering
    \includegraphics[width=0.6\linewidth]{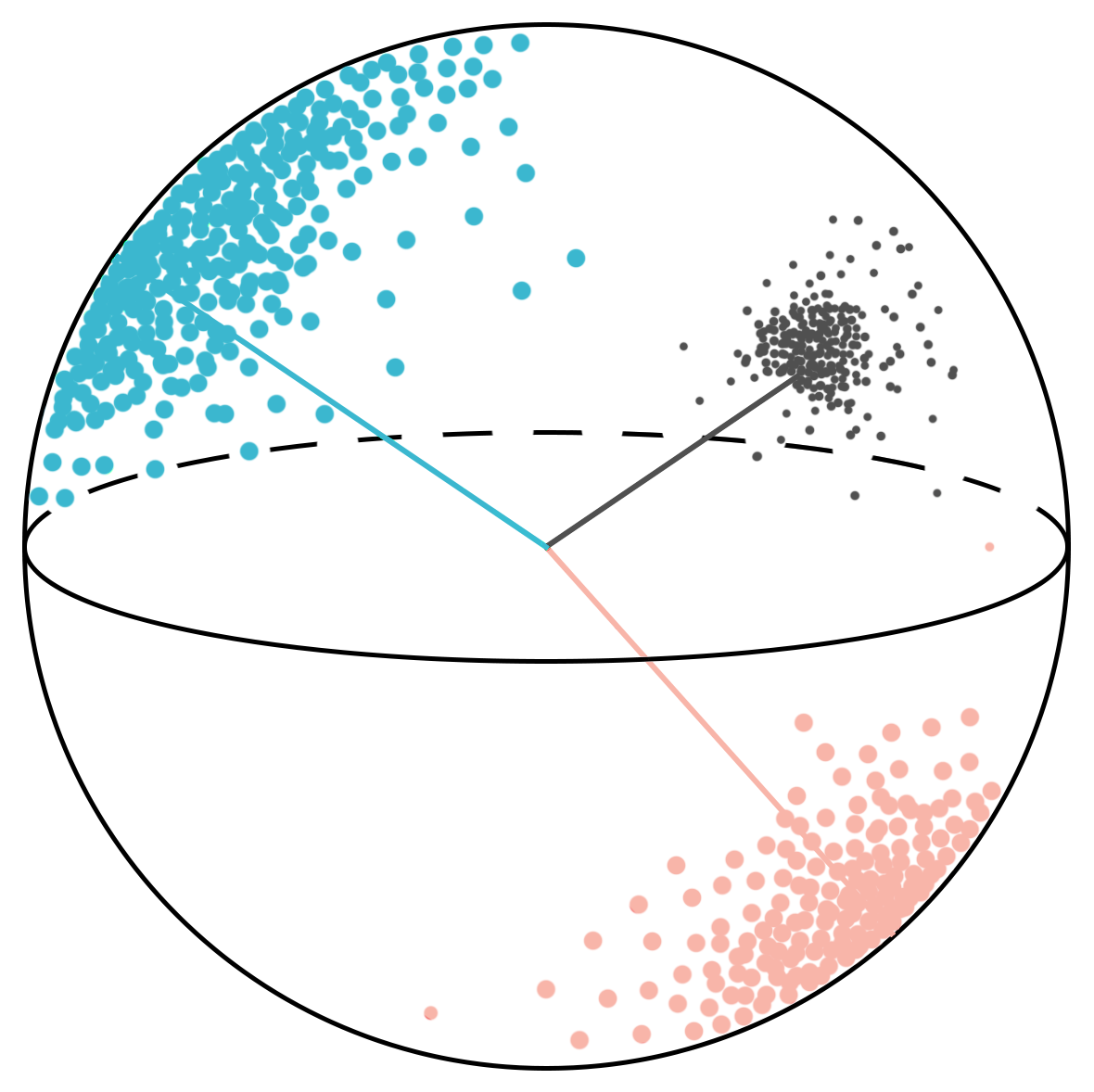}
    \caption{Clustering of data points on the unit hypersphere using the vMF distribution.}
    \label{fig:model_architecture}
\end{figure}

In this paper, we model latent embeddings with the von Mises–Fisher (vMF) distribution. vMF is analogous to a Gaussian distribution on the surface of the sphere $\mathcal{S}^{M-1}\subset\mathbb{R}^M$, where data points are unit vectors. Specifically, for $x\in\mathcal{S}^{M-1}$,
\begin{equation}
    q(x\mid \kappa,\mu)=C_M(\kappa)\,\exp\!\big(\kappa\,\mu^\top x\big),
\end{equation}
with normalizing constant
\begin{equation}
    C_M(\kappa)=\frac{\kappa^{M/2-1}}{(2\pi)^{M/2}\,I_{M/2-1}(\kappa)},
\end{equation}
where $I_{\nu}(\cdot)$ is the modified Bessel function of the first kind of order $\nu$.

Figure \ref{fig:model_architecture} illustrates three vMF clusters on $\mathcal{S}^2$. The vMF distribution models directional data on the hypersphere; the concentration $\kappa$ controls how tightly points cluster around the mean direction $\mu$ (larger $\kappa \to$ tighter clusters).

Intuitively, we map raw time series into an embedding space such that each class follows a vMF distribution (i.e., forms a cluster on the hypersphere). This serves two goals for deep hashing. First, the hypersphere preserves similarity before and after discretization \cite{hoe2021one}. Additionally, vMF clustering improves class compactness in the embedding space, enhancing hashing performance and suggesting margin-based modeling approaches.

\subsection{Overall Learning Framework}

\begin{figure}[htbp]
    \centering
    \includegraphics[width=1\linewidth]{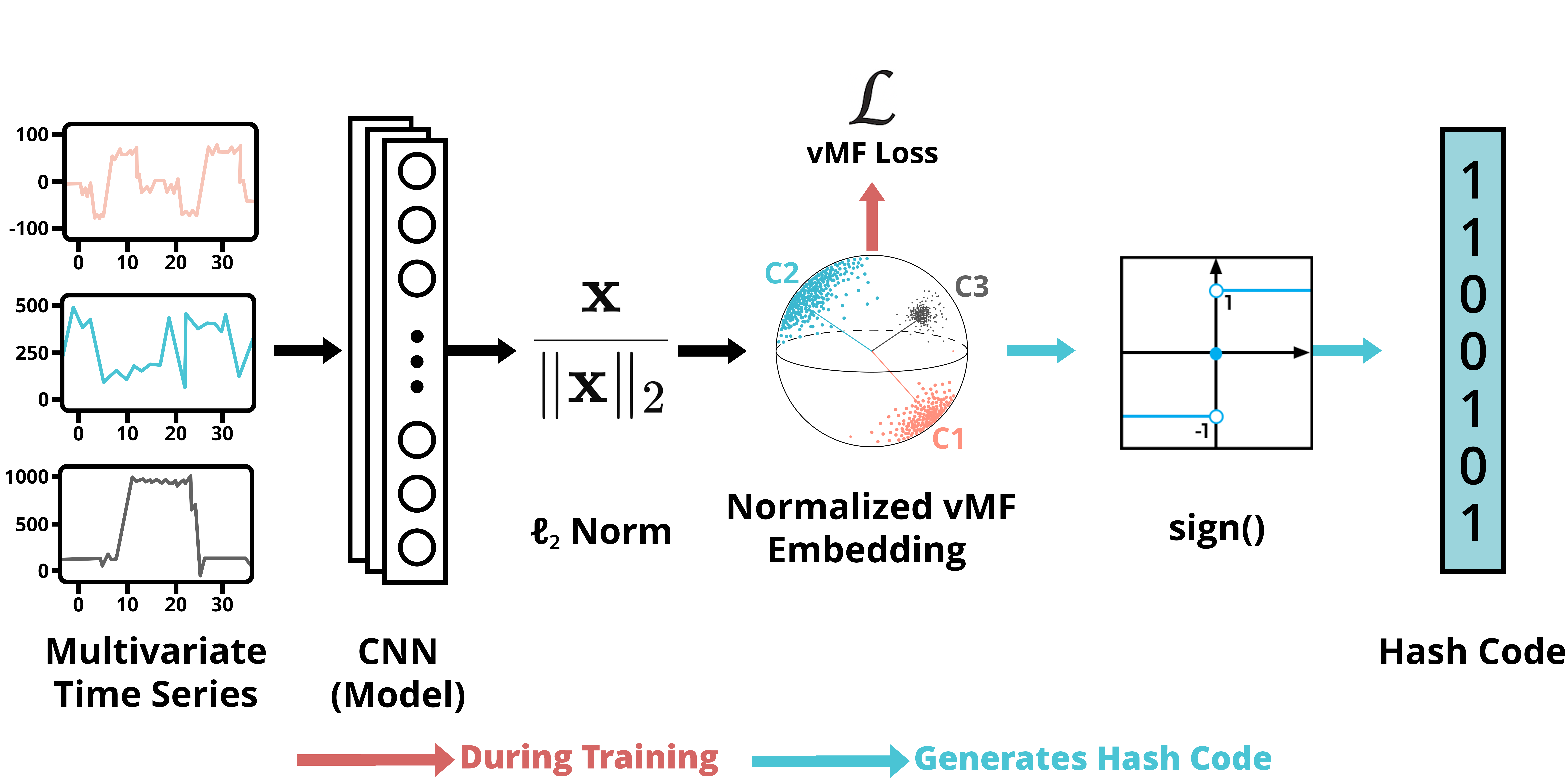}
    \caption{Proposed Framework}
    \label{fig:model_framework}
\end{figure}

Our proposed overall learning framework is illustrated in Figure \ref{fig:model_framework}. The black arrow indicates shared components of the pipeline, the red arrow shows operations performed during training, and the blue arrow indicates functions executed during the generation of the hash code. The process starts with the raw time series data passing through a 1D convolutional neural network encoder responsible for extracting high-level features from the raw input. After feature extraction, the vectors are normalized to unit length to ensure they lie on the unit hypersphere, preparing them for clustering using the vMF distribution. The normalized features are then fed into a fully connected layer that is tailored to work with the vMF distribution. The proposed loss function is applied by modeling the clustering performance via the log-likelihood probability. In the testing phase, the obtained embeddings are binarized using the sign function, which outputs our binary hash code for each input time series, ideal for indexing and similarity search tasks. 

Specifically, given a multivariate time series dataset $\mathcal{X}$, the 1D CNN model $f$ transforms the raw time series into latent embedding space $\mathcal{Z} \in \mathbb{R}^M$ through:

\begin{equation}
    z_i = f(x_i)
\end{equation}

where $x_i \in \mathcal{X}$ and $z_i \in \mathcal{Z}$.

Next, we map $z_i$ into the unit hypersphere $\mathcal{S}^{M-1}$ via $\ell_2$ normalization to minimize information loss after binarization:
\[
z_i^{\text{norm}}=\frac{z_i}{\lVert z_i\rVert_2}, \quad z_i^{\text{norm}}\in\mathcal{S}^{M-1}.
\]
Finally, $z_i^{\text{norm}}$ is mapped to Hamming space via
\[
b_i=\mathrm{sign}(z_i^{\text{norm}}),\qquad b_i\in\{-1,+1\}^M.
\]
Because $\mathrm{sign}(\cdot)$ does not have a gradient, we optimize $f$ on the continuous $z_i^{\text{norm}}$. Our loss $\mathcal{L}_{\text{vMF}}$ is defined on $z_i^{\text{norm}}\in\mathcal{S}^{M-1}$ to keep the post-binarization loss minimal.

\subsection{Estimating $\mu$ and $\kappa$}

Next, we introduce the parameter estimation used by the loss. Given a real-valued embedding $z_i^{\text{norm}}$ obtained from the model, we first estimate the vMF distribution for each class $k$. Specifically, to estimate the mean direction vector $\mu_k$ for a given set of embedding features $z_{i,k}^{\text{norm}}$ of class $k$:

\begin{equation}
    \mu_k = \frac{r_k}{\lVert r_k\rVert_2}
\end{equation}

$\mu_k$ is the cluster centroid on the hypersphere formed by the samples that belong to class $k$. $N_k$ is defined as the number of samples for class $k$. Intuitively, $\mu_k$ is the average direction of all class-$k$ vectors on the hypersphere.

Next, we estimate $\kappa_k$. Unlike $\mu_k$, the Maximum Likelihood Estimation (MLE) for $\kappa_k$ has no closed form due to the normalizer. We adopt a standard moment-based approximation \cite{davidson2018hyperspherical}:

\begin{equation}
    \kappa_k = \frac{M\,\bar r_k-\bar r_k^{\,3}}{1-\bar r_k^{\,2}}
\end{equation}

where ${r}_k$ is the sum of the unit-norm vectors for all samples belonging to class $k$

\begin{equation}
    r_k = \sum_{i=1}^{N_k} z_{i,k}^{\text{norm}}
\end{equation}

where $z_{i,k}^{\text{norm}}$ is the feature vector for the $i$-th sample of class $k$.

$\bar{r}_k$ is the mean resultant length for class $k$

\begin{equation}
   \bar{r}_k = \frac{\lVert r_k \rVert_2}{N_k}
\end{equation}

where $\lVert r_k \rVert_2$ is the norm of the resultant vector $r_k$, and $N_k$ is the number of samples in class $k$.

For high concentration ($\bar r_k>0.9$), we optionally use a numerically stable approximation for $\kappa_k$:

\begin{equation}
    \kappa_k = -0.4 + 1.39 \times \bar{r}_k + \frac{0.43}{1-\bar{r}_k}
\end{equation}

\subsection{Proposed vMF Hashing Loss}

Next, we introduce our proposed deep hashing loss. The loss function aims at reducing intra-class variance and increasing inter-class distance, resulting in well-separated clusters on the hypersphere. Unlike Gaussian-based models, vMF clustering preserves angular structure and is naturally suited for spherical environments. This helps reduce information loss during binarization: embeddings close to the surface retain directional fidelity when passed through the sign function. As a result, binary codes remain semantically meaningful, enhancing hashing performance in similarity-based retrieval tasks.

Given estimated parameters $\mu_c, \kappa_c$ for each class $c = 1,\dots,C$, we compute the directional similarity between a normalized embedding $z_i^{\text{norm}}$ and each class's vMF distribution. The log-probability of an $M$-dimensional data point $z$ under a vMF distribution with location $\mu$ and scale $\kappa$ is given by:

\begin{equation}
    \log \mathrm{vMF}(z_i^{\text{norm}}\mid\mu_c,\kappa_c)= \mathcal{L}'(z_i^{\text{norm}},\kappa_c,\mu_c)\;+\;\log C_M(\kappa_c)
\end{equation}

where $\log C_M(\kappa_c)$ is the log-normalization constant given by:

\begin{equation}
\log C_M(\kappa_c)=\Big(\tfrac{M}{2}-1\Big)\log\kappa_c - \tfrac{M}{2}\log(2\pi) - \log I_{\frac{M}{2}-1}(\kappa_c)
\end{equation}

Here, $I_{\frac{M}{2} - 1}(\kappa)$ denotes the modified Bessel function of the first kind, and $M$ is the dimensionality of the latent space.

The unnormalized log-likelihood term is defined as:

\begin{equation}
    \mathcal{L}'(z_i^{\text{norm}}, \kappa_c, \mu_c) = \kappa_c \,\mu_c^\top z_i^{\text{norm}}
\end{equation}

To compute the loss, we normalize the log-likelihoods via softmax and apply cross-entropy to enforce class alignment:

Let $\mathbf{Z}$ be the matrix of normalized embeddings, and $\mathcal{L}'_{ik}$ be the unnormalized log-likelihood of sample $i$ under class $k$. We normalize log-likelihoods using softmax to obtain a probability distribution:

\begin{equation}
p_{ic} = \frac{\exp\!\big(\log \mathrm{vMF}(z_i^{\text{norm}}\mid \mu_c,\kappa_c)\big)}{\sum_{j=1}^C \exp\!\big(\log \mathrm{vMF}(z_i^{\text{norm}}\mid \mu_j,\kappa_j)\big)}
\end{equation}

The final loss is computed using cross-entropy between $p_i$ and the ground-truth label $y_i$:

\begin{equation}
    \mathcal{L}(\mathbf{Z}, \mathbf{y}) = -\sum_{i=1}^N \log p_{i, y_i}
\end{equation}

This loss encourages high log-likelihood for the correct class and penalizes incorrect ones, promoting cluster compactness while maintaining a hyperspherical structure.

\subsection{Adaptive Margin via $\kappa$ Downscaling}

In addition to the vMF-based loss, we introduce an implicit margin mechanism that improves class separability. Prior to computing the log-likelihood, we downscale the concentration parameter $\kappa$ for each class:

\begin{equation}
\kappa_c' = \frac{\kappa_c}{\alpha}
\label{eq:amp_margin}
\end{equation}

where $\alpha$ is a positive constant (typically 2 or greater). This downscaling softens the vMF distributions by reducing their concentration, effectively flattening each class’s density on the hypersphere. This encourages embeddings to align more confidently with their correct class centroid to achieve a high likelihood score, while discouraging alignment with incorrect centroids. By doing so, the model implicitly enforces greater angular separation between clusters. Samples closer to decision boundaries must be more confidently placed toward their correct class direction to avoid confusion during the softmax operation. As a result, clusters form with greater spacing between them, naturally enhancing semantic separability. This downscaling is applied during training only; at test time we use the unscaled parameters.

\section{Experiments}
In this section, we will evaluate the performance of the proposed work on various baseline datasets and approaches. For all of the experiments, we utilize a ResNet \cite{he2016deep} as a backbone for our deep hashing framework. During training, we use Adam \cite{kingma2014adam} as the optimizer across all the experiments. Throughout all the experiments, we use Google Colab Python3 with an NVIDIA T4 GPU and we tested hash code performance at code lengths of 16, 32, 64, and 128 bits. 

\subsection{Datasets}

Our experiments use three different datasets to evaluate our performance.

\begin{itemize}
    \item \textbf{CharacterTrajectories} \cite{dau2019ucr}: 2858 character samples collected via a WACOM tablet, capturing x, y positions and pen tip force, which has been processed for noise reduction and normalization.

    \item \textbf{Sussex-Huawei Locomotion} \cite{wang2018summary}: A versatile annotated dataset of models of locomotion and transportation of mobile users. Following previous settings \cite{song2018deep}, we use the first subject, and we remove activity types with insufficient data volume. 

    \item \textbf{Physical Activity Monitoring} \cite{reiss2012introducing}: 18 physical activities performed by 9 subjects wearing 3 inertial measurement units and a heart rate monitor.
    
\end{itemize}

\subsection{Baselines}

We compared our approach with three different baselines:

\begin{itemize}
    \item \textbf{GreedyHash} \cite{su2018greedy}:  A deep hashing loss function that aims at optimizing binary codes by optimizing latent embeddings.

    \item \textbf{DHN} \cite{zhu2016deep}: A deep hashing loss function that takes the quantization error into consideration. 

    \item \textbf{HashNet} \cite{cao2017hashnet}: A deep hashing loss function that utilizes pairwise similarity among samples belonging to the same classes.
\end{itemize}

\begin{figure*}[!t]
  \centering

   \begin{subfigure}[t]{0.3\textwidth}
    \includegraphics[width=\linewidth]{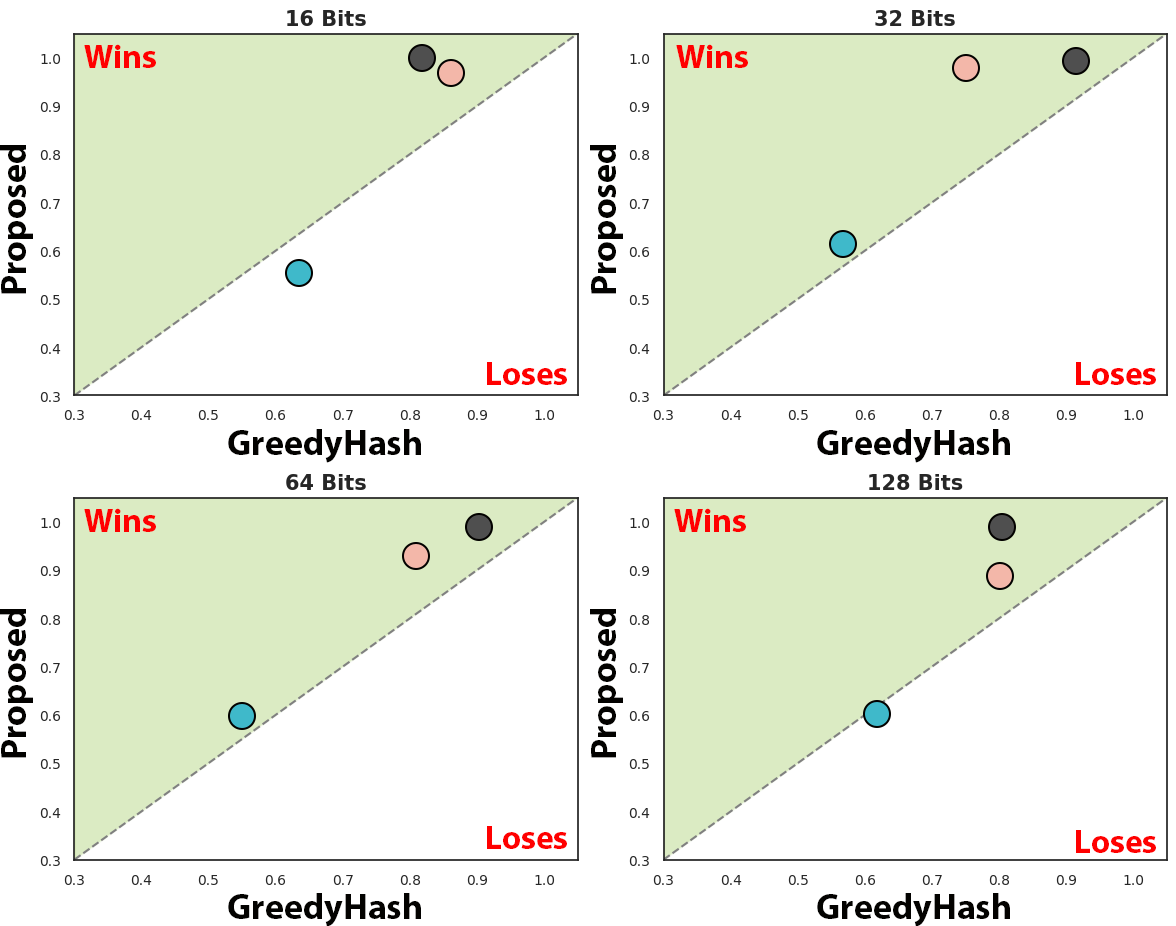}
    \caption{Proposed vs. GreedyHash}
  \end{subfigure}
  \hfill
  \begin{subfigure}[t]{0.3\textwidth}
    \includegraphics[width=\linewidth]{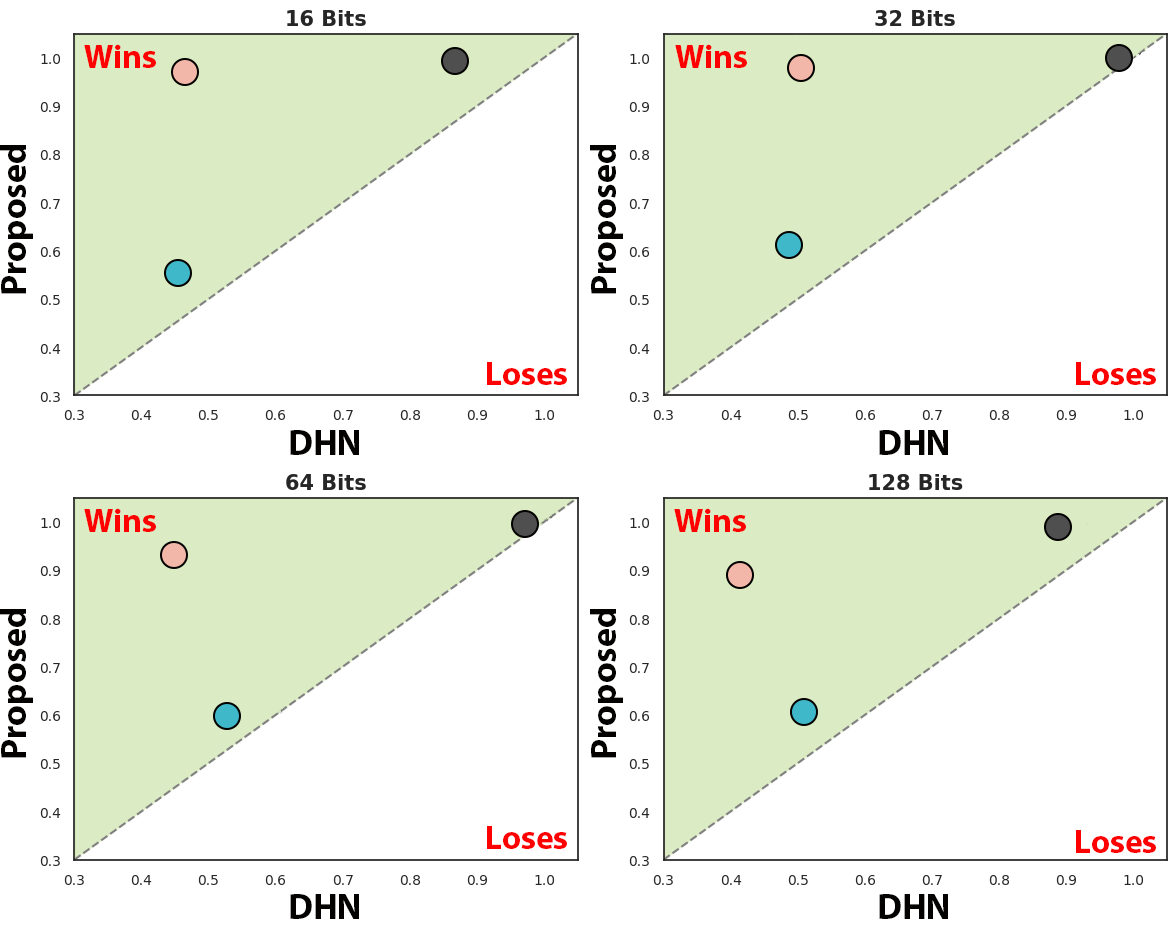}
    \caption{Proposed vs. DHN}
  \end{subfigure}
  \hfill
  \begin{subfigure}[t]{0.3\textwidth}
    \includegraphics[width=\linewidth]{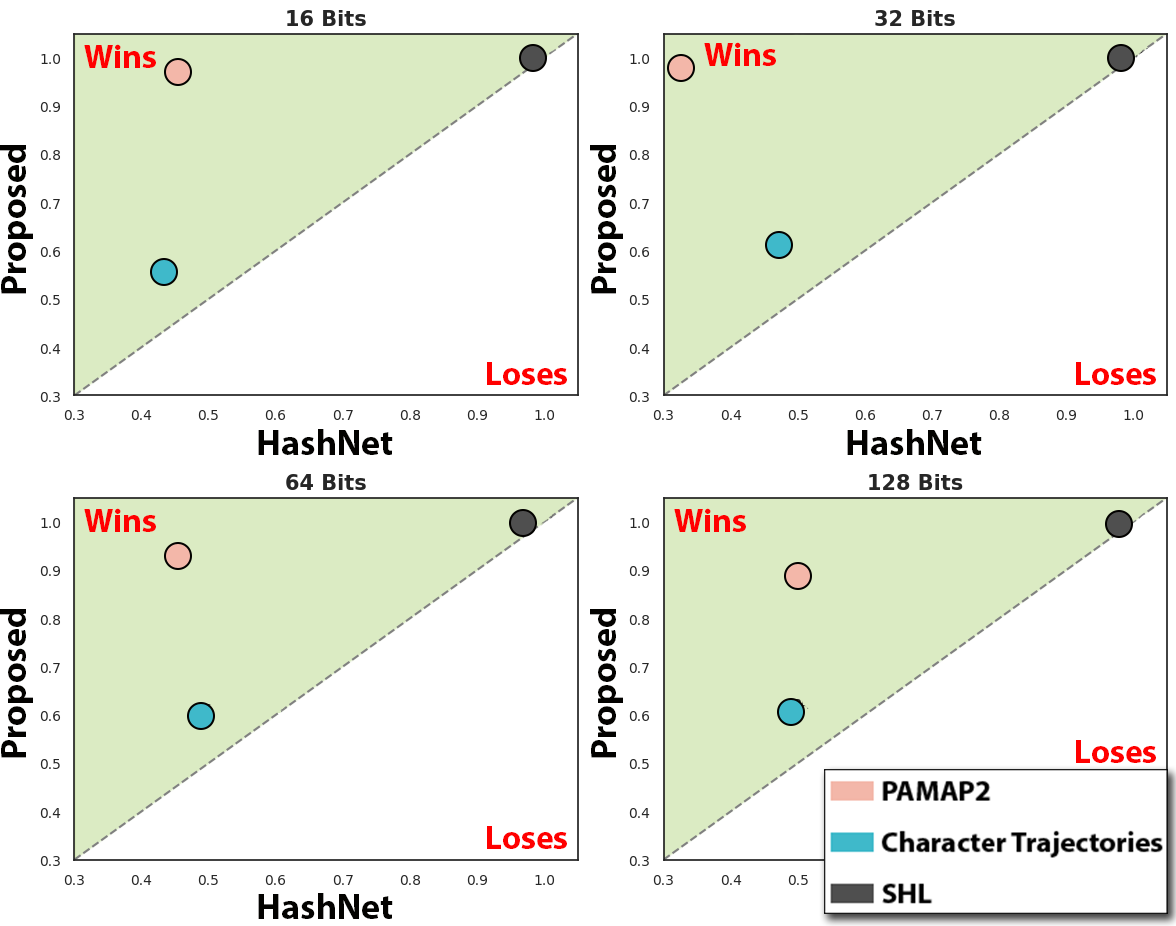}
    \caption{Proposed vs. HashNet}
  \end{subfigure}

  \caption{Comparison at different bit lengths. The green region indicates where our proposed model outperforms the baselines (GreedyHash, DHN, HashNet), while the white region indicates where it underperforms.}
  \label{fig:method_comparison}
\end{figure*}

\subsection{Evaluation Criteria}

We use the mean average precision (mAP) metric, a standard measure in hashing-based retrieval tasks. Given a set of query time series, a set of retrieval candidates, and their corresponding labels, mAP quantifies the quality of the top-$R$ retrieved samples for each query based on label relevance.

In the experiments, we use the top-$R$ nearest samples under the Hamming distance of the binary codes to retrieve the time series.  Given the labels $\{y_i\}$ for the top-$R$ retrieved time series, the average precision (AP) is computed as:

\[
\text{AP} = \frac{1}{R} \sum_{i=1}^{R} P(i)\,\mathbb{I}(y_i=1)
\]

where $P(i)$ is the precision at rank $i$, $\mathbb{I}(\cdot)$ denotes the indicator function (equal to $1$ if the item at rank $i$ is relevant, i.e., $y_i=1$, and $0$ otherwise), and $R$ is the retrieval cutoff. 

The final mAP is the mean of AP over all queries, where $n_q$ is the number of queries:
\[
\text{mAP} = \frac{1}{n_q} \sum_{j=1}^{n_q} \text{AP}_j
\]

To improve evaluation reliability, we randomly split the test hash codes into query and database sets. All results are averaged across five runs to account for randomness in hashing and label distributions.

\begin{table}[t]
\renewcommand{\arraystretch}{1.2}
\centering
\resizebox{\columnwidth}{!}{%
\begin{tabular}{lcccc}
\toprule
\multicolumn{5}{c}{\textbf{16 Bits}} \\
\midrule
Dataset & GreedyHash & DHN & HashNet & Proposed \\
\midrule
CharacterTrajectories & 0.856 & 0.465 & 0.455 & \textbf{0.972} \\
PAMAP2                 & \textbf{0.635} & 0.453 & 0.433 & 0.555 \\
SHL                    & 0.822 & 0.867 & 0.985 & \textbf{0.996} \\
\midrule
\multicolumn{5}{c}{\textbf{32 Bits}} \\
\midrule
CharacterTrajectories & 0.751 & 0.505 & 0.325 & \textbf{0.980} \\
PAMAP2                 & 0.567 & 0.487 & 0.471 & \textbf{0.613} \\
SHL                    & 0.913 & 0.980 & 0.985 & \textbf{0.996} \\
\midrule
\multicolumn{5}{c}{\textbf{64 Bits}} \\
\midrule
CharacterTrajectories & 0.807 & 0.449 & 0.455 & \textbf{0.932} \\
PAMAP2                 & 0.550 & 0.528 & 0.489 & \textbf{0.599} \\
SHL                    & 0.901 & 0.975 & 0.971 & \textbf{0.992} \\
\midrule
\multicolumn{5}{c}{\textbf{128 Bits}} \\
\midrule
CharacterTrajectories & 0.802 & 0.414 & 0.501 & \textbf{0.890} \\
PAMAP2                 & \textbf{0.614} & 0.508 & 0.488 & 0.608 \\
SHL                    & 0.803 & 0.888 & 0.980 & \textbf{0.992} \\
\bottomrule
\end{tabular}%
}
\caption{mean Average Precision (mAP) at Varying Code Lengths Across Three Datasets}
\label{tab:bitwise_map}
\end{table}

\begin{table}[t]
\renewcommand{\arraystretch}{1.2}
\centering
\resizebox{\columnwidth}{!}{%
\begin{tabular}{lcccc}
\toprule
Dataset & GreedyHash & DHN & HashNet & Proposed \\
\midrule
CharacterTrajectories & 0.804 & 0.458 & 0.434 & \textbf{0.944} \\
PAMAP2                 & 0.592 & 0.494 & 0.470 & \textbf{0.594} \\
SHL                    & 0.860 & 0.928 & 0.980 & \textbf{0.994} \\
\bottomrule
\end{tabular}%
}
\caption{Average mAP Comparison Across All Bit Lengths}
\label{tab:average_map}
\end{table}
\subsection{Results}

\noindent\textbf{Performance Comparison}: Tables~\ref{tab:bitwise_map} and~\ref{tab:average_map} present the mAP performance of the proposed vMF-based hashing method against three baselines at bits from 16 to 128. Overall, our approach consistently achieves the highest accuracy across nearly all configurations. On the CharacterTrajectories dataset, the proposed method outperforms all baselines at every bit length, achieving a peak mAP of 0.980 at 32 bits. This highlights the model’s ability to preserve local temporal patterns in fine-grained motion data. For the PAMAP2 dataset, although GreedyHash slightly outperforms our method at 16 and 128 bits, our vMF approach surpasses both DHN and HashNet consistently, and reaches its best performance of 0.613 at 32 bits. This demonstrates its robustness even on noisier sensor datasets. In the SHL dataset, our method performs on par with or better than all baselines, attaining 0.996 mAP at both 16 and 32 bits. Notably, our model maintains high performance even as the code length increases, highlighting its scalability and generalization. As shown in Table~\ref{tab:average_map}, our method achieves the highest average performance across all datasets with a mean mAP of 0.944, compared to 0.804 (GreedyHash), 0.458 (DHN), and 0.434 (HashNet). Furthermore, Figure~\ref{fig:method_comparison} shows one-vs-one comparisons between the proposed method and the three aforementioned baseline hashing algorithms (GreedyHash, DHN, and HashNet) across four code lengths (16, 32, 64, and 128 bits) and three datasets (CharacterTrajectories, PAMAP2, and SHL) represented by three different points in the image. Points above the diagonal line indicate that our method achieved higher mAP scores. From the figure, visually speaking, our proposed method is significantly better than the baselines in all testing cases. This confirms the effectiveness of the proposed adaptive vMF formulation for learning discriminative and compact hash codes.

\noindent\textbf{Number of Bits vs. Performance}: Next, we discuss the performance change across different bit settings. Across every bit size, our method yields better average results than the baselines. Furthermore, the average mAP across all bit settings is relatively stable (0.842 vs. 0.863 vs. 0.841 vs. 0.83) with 32 bits having the best performance. This demonstrates that our method is effective but also scalable and reliable across different bit settings.

\section{Conclusion}
    In this paper, we introduce a deep hashing approach based on the von Mises–Fisher distribution for multivariate time series data. We use this approach to model the latent embedding space to maximize the margin between semantically different samples. Through experimentation, we found this approach successful as it closely competes with existing baselines and in some cases clearly outperforms them, implying significant potential for this approach. Overall, vMF shows consistent performance, with stronger results at shorter bit lengths. GreedyHash also shows strong performance, with the PAMAP2 dataset being better than vMF at different bit lengths. With fine-tuning, we might be able to improve our results for the PAMAP2 dataset.

\section{Acknowledgments}

This work is supported by the National Science Foundation (NSF) under Grant CNS-2318682 and IIS-2348480, as well as U.S. Department of Education under GAANN (Grant No. 5100001128). The project’s computing resources are supported by Google Cloud Platform (GCP) credits.

\bibliography{transformer}
\bibliographystyle{plain}
\end{document}